\documentclass{IEEEtran4PSCC}
\ifCLASSINFOpdf
   \usepackage[pdftex]{graphicx}
\else
   \usepackage[dvips]{graphicx}
\fi
\usepackage{caption}
\usepackage[cmex10]{amsmath}
\usepackage{amssymb}
\usepackage{steinmetz}
\usepackage[noend]{algpseudocode}
\usepackage[dvipsnames]{xcolor}
\usepackage{tcolorbox}
\tcbuselibrary{listings, skins, breakable}

\usepackage{float}
\usepackage{algorithm}
\usepackage{algorithmicx}
\usepackage{mathtools}
\usepackage{graphicx}
\usepackage{dblfloatfix}
\usepackage{subcaption}
\usepackage[caption=false,font=normalsize,labelfont=sf,textfont=sf]{subfig}
\makeatletter
\let\old@ps@headings\ps@headings
\let\old@ps@IEEEtitlepagestyle\ps@IEEEtitlepagestyle
\def\psccfooter#1{%
    \def\ps@headings{%
        \old@ps@headings%
        \def\@oddfoot{\strut\hfill#1\hfill\strut}%
        \def\@evenfoot{\strut\hfill#1\hfill\strut}%
    }%
    \def\ps@IEEEtitlepagestyle{%
        \old@ps@IEEEtitlepagestyle%
        \def\@oddfoot{\strut\hfill#1\hfill\strut}%
        \def\@evenfoot{\strut\hfill#1\hfill\strut}%
    }%
    \ps@headings%
}
\makeatother

\begin{document}
%
\title{Constraint-Informed Active Learning for End-to-End ACOPF Optimization Proxies}

\author{
\IEEEauthorblockN{Miao Li \\ Michael Klamkin \\ Pascal Van Hentenryck}
\IEEEauthorblockA{Georgia Institute of Technology, Atlanta, USA\\ 
\{mli746, mklamkin, pvh\}@gatech.edu}
\and
\IEEEauthorblockN{Russell Bent}
\IEEEauthorblockA{Los Alamos National Laboratory \\
Los Alamos, NM, USA \\ 
\{rbent\}@lanl.gov}
\and
\IEEEauthorblockN{Wenting Li}
\IEEEauthorblockA{University of Texas at Austin, TX, USA\\
wenting.li@austin.utexas.edu}
}

\maketitle

\begin{abstract}
This paper studies optimization proxies—machine learning (ML) models trained to efficiently predict optimal solutions for AC Optimal Power Flow (ACOPF) problems. While promising, optimization proxy performance heavily depends on training data quality. To address this limitation, this paper introduces a novel active sampling framework for ACOPF optimization proxies designed to generate realistic and diverse training data. The framework actively explores varied, flexible problem specifications reflecting plausible operational realities. More importantly, the approach uses optimization-specific quantities (active constraint sets) 
that better capture the salient features of an ACOPF that lead to the optimal solution.
Numerical results show superior generalization over existing sampling methods with an equivalent training budget, significantly advancing the state-of-practice for trustworthy ACOPF optimization proxies.
\end{abstract}

\section{Introduction}
The security and economy of modern power grids hinge on solving optimal power flow (OPF) problems. In the ideal situation, grid planners and operators would rely on the alternating current (ACOPF) variation of the problem. Unfortunately, the non-convex and non-linear nature of the ACOPF formulation renders the use of conventional solvers computationally intractable for real-time applications, forcing operators to rely on simpler models. To address this issue, a growing line of research has explored the use of Deep Neural Networks (DNNs) as optimization proxies for ACOPF \cite{rosemberg2024learning,qiu2025dual, park2023self}, which can produce superior solutions to simpler problems with the benefit of millisecond inference speeds.
Within this paradigm, ensuring the robustness of the learning-based proxy emerges as a central challenge. 

Unlike traditional solvers, the performance of data-driven models is contingent on the breadth of its training data. While training instances can be generated by applying simulated real-time perturbations to forecasts \cite{chen2022learning}, it is often computationally impractical to cover or predetermine every possible operational scenario. Thus, a model trained on a limited dataset may not be reliable when presented with unforeseen conditions, yielding large deviations in legitimate realizations that the system must handle — and, if the proxy is inaccurate, the model directly undermines reliability of the predicted solution. 
To address this limitation, this paper introduces a novel active learning (AL) framework for optimization proxies that promotes sample diversity based on features of optimization solutions. 
The approach builds on the work of \cite{klamkin2024bucketized}, which introduced input space partitioning (buckets) for ACOPF. These partitions are defined by factors like load perturbation and were used to guide an AL sampling procedure to meaningful examples to include in training.
The approach of this paper is motivated by the observation that 
while this AL demonstrated strong average performance, it did not capture the structural properties of the underlying ACOPF optimization problem, thereby resulting in large outlier prediction errors.

Instead, the paper interprets the spatial representation of the ACOPF through its active constraint sets, which describe how the solution polytope evolves in response to changes in the ACOPF's input parameters. The approach identifies shifts in these active sets, in particular those shifts that \emph{yield significant changes in solution sensitivity to input parameters}, which are difficult for a DNN to predict without representative examples.
The effectiveness of this framework is validated through a rigorous evaluation focused on performance of worst-case errors to confirm the AL's sample distribution is both representative and robust. 
In short, this paper makes the following key contributions:
\begin{enumerate}
\item 
The introduction of novel features based on active constraint sets in optimization problems like ACOPF.
\item Integration of active constraint set features into AL which significantly reduces tail prediction errors in ACOPF optimization proxies. 
\item Extends AL for ACOPF optimization proxies to handle diverse load profile distributions that model multiple operational scenarios.
\item Generalizes the AL partitioning scheme of 
\cite{klamkin2024bucketized} to a larger class of feature spaces and training scenarios.
\end{enumerate}
\vspace{-0.2cm}

\section{Related work}
Most active learning literature centers on query-focused methods that label data based on acquisition scores from the learning model. Foundational strategies score individual samples based on model uncertainty \cite{wu2018pool}\cite{sauer2023active}\cite{park2020robust}. Other work acquires batches of samples that are simultaneously informative and diverse based on model-derived metrics. These techniques include, for instance, using gradient-based sampling to find varied sets of high-uncertainty points \cite{ash2019deep} and using information-theoretic principles to maximize the collective information gain \cite{kirsch2019batchbald}.

In contrast, a smaller class of methods uses awareness of the global data distribution. One example is the core-set approach \cite{sener2017active}, which prioritizes samples that are holistically representative of the entire unlabeled pool. A limitation of the core-set and similar methods is that they do not directly integrate information about the model's performance in specific regions of the input space. Therefore, they overlook how the model's intrinsic architecture responds to different features in the input space, leading to uneven performance.
To address this gap, a specialized method for ACOPF proxy active learning was developed in \cite{klamkin2024bucketized}, which considered both the input space distribution and model performance. This technique partitions (referred to as buckets) the input space based on load perturbations, a domain-specific feature of ACOPF. The approach then uses a held-out validation set to assess the model's performance within each bucket and guide the selection of additional training points for labeling. The held-out validation set directly measures performance and was more reliable than sampling based on model-derived uncertainty estimates. This type of evaluation-centric approach was also successfully employed in several other prominent active learning frameworks \cite{liu2021influence, ding2024learning, killamsetty2021glister}.

An important aspect of the partitioning scheme in \cite{klamkin2024bucketized} is the assumption that all potential samples inside a bucket are equally informative. In practice, a bucket could include samples with a variety of features that influence ACOPF solution prediction, thereby requiring very fine partitioning to acquire useful samples. Unfortunately, such resolution introduces computational complexity (performance assessment of each bucket) and the risk of redundant bucket creation, where many buckets are similarly informative. 
In contrast, this paper focuses on the underlying geometry of ACOPF optimization and the structures that \emph{directly influence the optimal solution}. It uses active constraint sets as a structural abstraction of this underlying geometry to guide sampling within the buckets of \cite{klamkin2024bucketized}, thereby reducing intra-bucket performance variance.

\section{Modeling Background}
This section presents the background of the ACOPF formulation, the general active learning framework, and the data preparation process. For notational simplicity, a machine learning feature is denoted by \( x \in \mathcal{X} \), and the corresponding label is denoted by  \( y \in \mathcal{Y} \). The true joint distribution is denoted by $p_z$, with ${z = (x,y)}$. 
\vspace{-0.1cm}
\subsection{ACOPF}
Model~\ref{eq:opf_system} presents the ACOPF formulation considered in this work. This nonlinear optimization problem minimizes the total cost of electricity generation, as defined in \eqref{eq:cost}, and follows the standard formulation in \cite{klamkin2025pglearn}. The primary decision variables include active and reactive power generation $S^{g} = p^{g} + iq^{g}$, power flows across the network's transmission lines in forward and reverse directions $S^{f} = p^{f} - iq^{f}$, $S^{t} = p^{t} - iq^{t}$; and the complex voltage, $V$, where $v = |V|$ denotes the voltage magnitude and $\theta = \phase V$ the angle in the complex plane. The model is constrained by the laws of physics—including power balance at each bus, where the demand is denoted as $S^d = p^d+iq^d$, and nodal shunt conductance and susceptance are $Y^s = g^{s}+ ib^{s}$ (constraint \ref{eq:power_balance}). Power flow is modeled with Ohm’s law  through constraints (\ref{eq:pf_from}--\ref{eq:pt_from}), where $Y$ is the complex admittance of the line  and $W =VV^*$. Engineering limits are enforced throughout the model: thermal line capacities (constraints \ref{eq:thermal_capacity_from}--\ref{eq:thermal_capacity_to}), angle-difference bounds on transmission lines (constraints \eqref{eq:angle_diff_bounds}), generator output limits (constraints \eqref{eq:pg_bounds}), voltage magnitude bounds (constraints \eqref{eq:voltage_bounds}), and power-flow limits (constraints \ref{eq:pf_from_bounds}--\ref{eq:pf_to_bounds}).
\vspace{-0.9cm}
\begin{center}
\noindent\rule{0.5\textwidth}{0.1pt} \\ 
\textbf{Model 1 AC Optimal Power Flow (AC-OPF)}\\
\noindent\rule{0.5\textwidth}{0.1pt} \\ 
\vspace{-0.5cm}
{  \setlength{\abovedisplayskip}{3pt}
  \setlength{\belowdisplayskip}{0pt}
\begin{subequations} \label{eq:opf_system}
\begin{align}
&\min_{S^g, S^f, S^t, \mathbf{v}, \boldsymbol{\theta}
} \quad  \sum_{i \in \mathcal{G}} c_i p_i^g  \label{eq:cost} \\
&\text{s.t.} \quad
 \sum_{j \in \mathcal{G}_i} S_{j}^g - \sum_{j \in \mathcal{L}_i} S_{j}^d -  Y^s v_i^2 = \sum_{e \in \mathcal{E}_i^f} S_e^f + \sum_{e \in \mathcal{E}_i^t} S_e^t \quad \forall i \in \mathcal{N}\label{eq:power_balance} \\
& S_e^f = Y_e^{ff} v_i^2 + Y_e^{ft*}W_{ij} \quad e \in \mathcal{E} \label{eq:pf_from} \\
& S_e^t = Y_e^{tt} v_i^2 + Y_e^{tf*}W_{ij} \quad e \in \mathcal{E} \label{eq:pt_from} \\
& S^{f*}_e S^f_e \leq \bar{S}_e^2  \quad  \forall e \in \mathcal{E}\label{eq:thermal_capacity_from} \\
&  S^{t*}_e S^t_e \leq \bar{S}_e^2 \quad \forall e \in \mathcal{E} \label{eq:thermal_capacity_to} \\
& \Delta \underline{\theta}_e \leq \theta_i - \theta_j \leq \Delta \bar{\theta}_e  \quad \forall e \in \mathcal{E}\label{eq:angle_diff_bounds} \\
& \theta_{\text{ref}} = 0\label{eq:ref_angle} \\
& \underline{S}_i^g \leq S_i^g \leq \bar{S}_i^g \quad\forall i \in \mathcal{G}\label{eq:pg_bounds} \\
& \underline{v}_i \leq v_i \leq \bar{v}_i \quad \forall i \in \mathcal{N}\label{eq:voltage_bounds} \\
& -\bar{S}_e \leq S_e^f \leq \bar{S}_e  \quad \forall e \in \mathcal{E}\label{eq:pf_from_bounds} \\
& -\bar{S}_e \leq S_e^t \leq \bar{S}_e  \quad \forall e \in \mathcal{E}\label{eq:pf_to_bounds} 
\end{align}
\end{subequations}}
\noindent\rule{0.5\textwidth}{0.1pt}
\end{center}
\vspace{-0.2cm}
In this formulation several sets are defined, including $\mathcal{N}$ for the set of buses. Each bus $i \in \mathcal{N}$ is associated with generator and load sets, denoted by $\mathcal{G}_i$ and $\mathcal{L}_i$, respectively. The set of branches, including transmission lines and transformers, is denoted by $\mathcal{E}$. Each $e \in \mathcal{E}$ connects buses $i$ and $j$, with $(i,j)$ specifying its direction. To account for parallel branches sharing the same endpoints, branches are identified directly by their terminal buses, i.e., $e = (i,j) \in \mathcal{E}$. The sets of branches leaving and entering bus $i$ are represented by $\mathcal{E}_i$ and $\mathcal{E}_i^R$, respectively. 
\subsection{Active learning}
Active sampling trains high-performing models with minimal labeling by iteratively selecting a small, maximally informative, and representative subset $\mathcal{D}^+\subseteq\mathcal{D}_U$, where $\mathcal{D}_U$ is a large pool of data whose labels are initially unknown. Data from $\mathcal{D}_U$ is then added to the labeled training set $\mathcal{D}$ to progressively expand it in each round. Input features $x$ from $(x,y) \in \mathcal{D}_U$ are drawn from the marginal $p_x$ of an unknown joint distribution $(x,y)\sim p$. Labels, $y$, are acquired only for the queried feature $x$ in $\mathcal{D}$. By precisely selecting labels within the feature space, the goal is to reduce epistemic errors and minimize the cost of labeling. This paper considers the commonly used batch-based active learning problem formulation. Here, the optimal query strategy selects a batch $\mathcal{D}^{+}$ of $\beta$ new instances (i.e., $|\mathcal{D}^{+}|=\beta$) per query round. This selection guides the search for the most accurate model $h$ from the function space of candidate models, $\mathcal{H}$. This ideal selection process, consisting of input features $\mathbf{x}$, output labels $\mathbf{y}$, a selected metric $M$ and a loss function $\mathbb{L}(h(\mathbf{x}), \mathbf{y})$, is generally formulated as the following {optimization problem}:
{\setlength{\abovedisplayskip}{4pt}
  \setlength{\belowdisplayskip}{3pt}\begin{equation} \label{eq:al_generic_formulation}
\begin{split}
\mathcal{D}^+ \coloneq & \underset{\mathcal{D'} \subseteq \mathcal{D}_U, |\mathcal{D}'|=\beta}  {\operatorname{argmin}}  \quad M( \mathbb{L}(h^*({x}), {y})) \\
 \text{s.t.}  \quad  &h^* \coloneq \underset{h\in  \mathcal{H} }{\operatorname{argmin}} \quad \sum_{({x}, {y}) \in \mathcal{D}\bigcup\mathcal{D}'} \mathbb{L}(h({x}), {y})
\end{split}
\end{equation}}
where, for a choice of dataset $\mathcal{D}'$, the inner optimization searches for the optimal mapping $h^*$, that minimizes loss $\mathbb{L}$ on dataset $\mathcal{D} \bigcup \mathcal{D}'$. Then, the outer optimization searches for the dataset $\mathcal{D}'$, whose optimal model, $h^*$, delivers the best performance based on metric $M$. The most standard choice of $M$ is the expected value with respect to the true distribution $(x, y) \sim p$, {\setlength{\abovedisplayskip}{1pt}
  \setlength{\belowdisplayskip}{3pt}\begin{equation*}
    \mathcal{D}^+ \coloneq \underset{\mathcal{D'} \subseteq \mathcal{D}_U, |\mathcal{D}'|=\beta}  {\operatorname{argmin}}  \quad \mathbb{E}_{(x, y) \sim p} \left[ \mathbb{L}(h^*({x}), {y}) \right]. 
\end{equation*}}
To ensure the constructed optimization proxy is reliable and robust, this work evaluates its tail-end performance (e.g., the 90th error percentile) alongside its mean performance. Section~\ref{sec:exp} presents results for both types of metrics
Note that directly solving bilevel optimization problems is often intractable, heuristic methods are commonly employed as a practical alternative.
\subsection{Data Generation}\label{subsec:data}
The development of an ACOPF optimization proxy requires large-scale datasets of solved problem instances, constructed by sampling diverse operating conditions and solving each instance to optimality. The resulting labeled dataset maps ACOPF inputs to corresponding ACOPF solutions. In practice, these input features consist of active and reactive load profiles, which may differ substantially across operating conditions and over time \cite{chen2022learning, chen2018developing}.
A major challenge is the scarcity of publicly available datasets with sufficiently diverse load profiles. Most datasets provide a single static nominal point $x_0 = S^d_0,$ with other parameters (e.g., generator bounds $\bar{S}^g, \underline{S}^g$) kept fixed across instances. As emphasized in prior work \cite{gayme2012optimal}, however, ACOPF solutions must be evaluated over a broad range of load conditions, and reliance on a single nominal point is sometimes insufficient to capture this variability.  
To address this limitation, Algorithm \ref{algo:data generation} generates synthetic load profiles. Assuming, without loss of generality, that the nominal case can serve as a representative baseline for peak demand, and scaling it with stochastic multipliers to produce low, mid, and high activity levels, generating $N$ samples at each level. This approach provides a practical and sufficiently general scheme for creating a diverse set of test instances. Line 5 specifies, for each bus $i \in \mathcal{N}$, the active demand at level $t\in\mathcal{T}$ is sampled as $ S^d_{t,i} = m_{t,i} \, S^d_{0,i},$ where $m_{t,i}$ is drawn from a distribution $\mathcal{M}_t$ conditioned on a demand regime:
{\setlength{\abovedisplayskip}{2pt}
  \setlength{\belowdisplayskip}{1pt}\[
\mathcal{M}_t \triangleq
\begin{cases}
U[0.55, 0.7) & \text{for } t = t_\text{low}, \\
U[0.7, 0.85) & \text{for } t = t_\text{mid}, \\
U[0.85, 1.0) & \text{for } t = t_\text{high}.
\end{cases}
\]}
\vspace{-0.2cm}
\begin{algorithm}[H]
    \caption{ACOPF Load Perturbation}
    \begin{algorithmic}[1]
        \State \textbf{Input:} Nominal sample $x_0$, regional load variation distribution $\mathcal{Q}_r$, individual noise variation distribution $\mathcal{Q}_i$, demand profiles $\mathcal{M}_t$ and data set size $N$ for each time point.
        \State Initialize $\mathcal{D} \leftarrow \emptyset$. 
        \For {$t \in \mathcal{T} $}
            \For{$n = 1 \dots N$}
            \State $m_{t,i}^n \sim \mathcal{M}_t$, where $i\in \mathcal{N}$
            \State $d_t^n \sim \mathcal{Q}_r$, $\epsilon_t^n \sim \mathcal{Q}_i$
            \State $x_t^n \leftarrow b_t^n \cdot m_{t}^n \circ x_0 \circ \epsilon_t^n$
            \State $y_t^n \leftarrow \text{ACOPF}(x_t^n)$
            \State$\mathcal{D} \leftarrow \mathcal{D} \cup \{(x_t^n, y_t^n)\}$
            \EndFor
        \EndFor \\
\Return{ $\mathcal{D}$} 
    \end{algorithmic}
    \label{algo:data generation}
\end{algorithm}
\vspace{-0.1cm}
For the purposes of this paper, further data augmentation is constructed around each of the previously generated nominal operating points $m_{t,i}^n$, ensuring that the training distribution spans a diverse set of conditions. The augmentation includes both regional $\mathcal{Q}_r$ and individual $\mathcal{Q}_i$ perturbations, following standard data augmentation practices in OPF learning \cite{klamkin2024bucketized, pan2022deepopf, owerko2020optimal, dong2020smart, zamzam2020learning, ferrando2023physics, baker2020learning, guha2019machine} as detailed in lines 6-7. In Lines 8–9, the corresponding solutions are generated and aggregated.

\section{Active learning for ACOPF}\label{section:algorithm}
The design of the algorithm centers on two objectives--ensuring sufficient coverage of the entire input space (diversity) and selecting samples that most improve the model (informativeness). The first goal is achieved through bucketization (Section \ref{subsec:Bucketization}), while the second is accomplished using an active set filter focusing on the worst-performing points within each bucket (Section \ref{subsec:Active Set Filter}). 
The high level structure of this Algorithm is  given in Algorithm \ref{Active Learning Loop}.
The algorithm begins with a definition of the data and associated hyperparameters (line 1).  The data is split into three sets: an initial set of training data ($\mathcal{D}_0$), a set of data for evaluating the performance of the ACOPF proxy ($\mathcal{D}_V$), and a set of unlabeled data that is available to add to the training set ($\mathcal{D}_p$). The hyperparameters include the number of buckets ($k$), the labeling budget per round ($\beta$), and the number of query rounds ($R$). Line 2 partitions the input space into a set of $k$ buckets, $\mathcal{B}$. Here, the notation $b(d)$ is used to denote the data from set $d$ contained in bucket $b$. The details of how the buckets are created is described in \ref{subsec:Bucketization}.

\begin{algorithm}[H]
    \caption{Active learning for ACOPF with Active set filters}
    \label{Active Learning Loop}
    \begin{algorithmic}[1]
        \State \textbf{Inputs:} Initial training data set $\mathcal{D}_0$; Bucket-validation set $\mathcal{D}_V$ and unlabeled data set $\mathcal{D}_p$. Architecture of DNNs $\hat{h}$ and $\hat{h}_a$; number of query rounds $R$ and query size $\beta$; hyperparameter $k$ for the number of buckets.
        \State \textbf{Create} a set of $k$ buckets, $\mathcal{B}$.
        \State $\mathcal{D}\leftarrow\mathcal{D}_0$;
        \For{$r=1:R$}
            \State $\triangleright$ Train $\hat{h}$ and $\hat{h}_a$ on $\mathcal{D}$ until convergence.
            \State $\triangleright$ Active Sampling:
            \State \quad Compute the bucket-wise acquisition scores:
            \State \quad \quad $S = \{s_b : s_b \leftarrow \alpha(b(\mathcal{D}_v), \hat{h}), \forall b \in \mathcal{B}\}$,
            \State \quad \quad $\mathcal{N} = \{n_b : n_b \leftarrow \eta(s_b, b), \forall b \in \mathcal{B}\}$, 
            \State \quad \quad with $\sum n_b = \beta$.
            \For{$b \in \mathcal{B}$}:
            \State  Construct  $\mathcal{X}_b = \sigma(b({\cal D}_v))$ with $|\mathcal{X}_b|=n_b$. 
            \State Define for $ x \in \mathcal{X}_b$, \State\quad$\Upsilon_{x}=\{x'\in \mathcal{D}_p: \mathcal{A}(x')=\mathcal{A}(x)\}$. 
            \State Select samples from $b(\mathcal{D}_{p})$ to form $b(\mathcal{D}^{+})$:
            \State \quad {  \setlength{\abovedisplayskip}{0pt}
  \setlength{\belowdisplayskip}{0pt}$\mathcal{D}^{+} = \bigcup_{x \in \mathcal{X}_b} \left\{ \underset{\upsilon \in \Upsilon_x}{\text{argmin}} \, \delta(x, \upsilon) \right\}$}.
            \State Update the training set and unlabeled pool:
                        \State \quad \quad $\mathcal{D} = \mathcal{D} \cup \mathcal{D}^{+}$ and $\mathcal{D}_{p} = \mathcal{D}_{p} \setminus \mathcal{D}^{+}$.
            \EndFor
        \EndFor
        \State \textbf{Outputs:} The final ACOPF proxy $\hat{h}$.
    \end{algorithmic}
\end{algorithm}
\vspace{-0.4cm}

Line 3 then initializes the active sample set, $\mathcal{D}$, with $\mathcal{D}_0$. Lines 4-20 form the bulk of the active sampling by iteratively augmenting $\mathcal{D}$ with additional samples ($R$ times).  Within this loop, line 5 trains the proxy model, $\hat{h}$ and trains a model for predicting active sets, $\hat{h}_a$, as is described in Sections \ref{subsec:proxy}.
Lines 8-10 then evaluate the performance of $\hat{h}$ on $\mathcal{D}_v$ as described in Section \ref{subsec:scoring}.

Next, lines 11-20 perform the active sampling from each bucket. 
More specifically, line 12 identifies points
$(\mathcal{X}_b)$ from $\mathcal{D}_v$ (a combination of poor performing and random points). Here, the notation $\sigma(d)$ is used to denote the selected combination from the set $d$, based on a selection function $\sigma$. The details are described in section \ref{subsec:Active Set Filter}. For any given $x \in \mathcal{X}_b$, defines the set of instances in $\mathcal{D}_p$ that share the same active set at optimality as $x$. Here, $\mathcal{A}(x)$ denotes the active constraint set corresponding to the optimal solution. Further details are provided in \ref{subsec:Active Learning Candidate Pool}. Line 16 then selects underrepresented points—quantified by $\delta$—with respect to active sets, to augment the training set (choices for $\delta$ are described in Section~\ref{subsec:Active Learning Selection}).

\subsection{Bucket Creation}\label{subsec:Bucketization}

Line 2 of Algorithm \ref{Active Learning Loop} divides the feature (data) space of $\mathcal{D}_v \bigcup\mathcal{D}_p$ into $k$ buckets, forming a set of buckets, $\mathcal{B}$. 
In principal, buckets defined on active-set combinations would be ideal, but is impractical as the number of active set combinations grows exponentially and many combinations are unrealizable or implausible, and identifying the active set of a given data without solving the optimization remains an open problem.
Instead, this paper builds on the load-perturbation buckets introduced by \cite{klamkin2024bucketized}, using a k-MEANS++ \cite{arthur2006k} bucketization approach, a computationally efficient approximation of the $k$-Determinantal Point Process ($k$-DPP)\cite{kulesza2011k}.

k-MEANS++ starts by selecting the first center randomly and then each subsequent center with probability proportional to its squared distance from the nearest chosen center, resulting in more stable clustering than standard k-means, which selects all centers randomly.

\subsection{Optimization Proxy Training}\label{subsec:proxy}

Line 5 iteratively trains two deep neural networks (DNN) each time the training data set, $\mathcal{D}$, is updated with active samples.  The first DNN is the ACOPF proxy,  \(\hat h:\mathcal{X}\to\mathcal{Y}\), which maps demands \(x=S^d\) to ACOPF solutions \(y=[S^g,v,\theta]\). This DNN is defined layer-wise for $l \in [1,L]$ by 
{\setlength{\abovedisplayskip}{3pt}
  \setlength{\belowdisplayskip}{3pt}
 \begin{align*}
 {\hat{h}}^0({x}) &= {x} \\
 {\hat{h}}^l({x}) &= \pi({W}^l{\hat{h}}^{l-1}({x}) + \mathbf{b}^l), \ \text{for } l\in [1,L-1]\\
 {\hat{h}}^L({x}) &=  {W}^L{\hat{h}}^{L-1}({x}) + \mathbf{b}^L
 \end{align*}}
\noindent with parameters \(\gamma=\{W^{l},b^{l}\}_{l=1}^{L}\), and activation function $\pi$, which is chosen as ReLU in this paper. $\gamma$ is learned using the empirical risk minimization function: $\gamma^{*}\coloneq \arg\min_{\gamma}\sum_{(x,y)\in\mathcal D}\mathcal L\big(\hat h(x),y\big)$, where {\setlength{\abovedisplayskip}{2pt}
  \setlength{\belowdisplayskip}{1pt}\begin{align}\label{eqn:training_loss}
    {\cal L}\big(\hat h(x),y\big) =\|y-\hat{h}(x)\|_{\ell_2}. 
\end{align}}

Next, as with creating buckets, active constraints present a major challenge for the selection of data to add to $\cal D$ as candidate data with the desired active constraints cannot be identified without conducting an optimization on points in ${\cal D}_p$. 
Instead, line 5 trains a second DNN predictor, \(\hat h_a:\mathcal{X}\to\mathcal{Y}_a\), which maps demands \(x=S^d\) to active constraints
\(a = [a^h, \ldots, a^r]\in\mathcal{Y}_a\), where each component of the output corresponds to binary indicators on inequality constraints defined in \eqref{eq:thermal_capacity_from}–\eqref{eq:angle_diff_bounds} and \eqref{eq:pg_bounds}–\eqref{eq:pf_to_bounds}. $\mathcal{Y}_a$ denotes the space of binary vectors, and its dimension is denoted by $K$.

Here, the DNN architecture is the same as $\hat{h}$ for the hidden layers $l\in[1,L-1]$, and the output layer for $\hat{h}_a$ converts the prediction to probability scores using the softmax function:
{\setlength{\abovedisplayskip}{2pt}
  \setlength{\belowdisplayskip}{3pt}\[
\tilde{a}_k = \frac{e^{\hat{h}_a^{L-1}(x)}}{\sum_{k=1}^{K} e^{\hat{h}_a^{L-1}(x)}}.
\]}
The binary cross entropy loss, 
{\setlength{\abovedisplayskip}{4pt}
  \setlength{\belowdisplayskip}{3pt}\begin{align*}
    \mathcal{L}_{as} = - \frac{1}{K} \sum_{k=1}^K \left[
    a_k \log\left( \tilde{a}_k \right) +
    (1 - a_k) \log\left( 1 - \tilde{a}_k \right)
\right], 
\end{align*}}
is used as the loss function for training $\hat{h}_a$. 

\subsection{Active Learning Scoring}\label{subsec:scoring}
Lines 8 and 9 then compute acquisition scores for each bucket based on the performance of $\hat h$ on the validation set, ${\cal D}_v$. Here, the acquisition function $\alpha$ models the urgency of acquiring new samples for each $ b\in\mathcal{B}$. This paper's acquisition function $\alpha$ adopts the Input-Loss-Gradient (IG) loss formulation of \cite{klamkin2024bucketized}:
{\setlength{\abovedisplayskip}{2pt}
  \setlength{\belowdisplayskip}{3pt}
\begin{align}\label{eqn:acquisition}
\alpha(b(D_v), \hat{h}) = \frac{1}{|b(D_v)|} \sum_{(\mathbf{x}_j, \mathbf{y}_j) \in b(D_v)} \left\lVert \nabla{\mathbf{x}_j} \mathcal{L}(\hat{h}(\mathbf{x}_j), \mathbf{y}_j) \right\rVert,
\end{align}}
Next, the distributor function $\eta(s_b, b)$ determines the number of samples $n_b$ to generate for each bucket, $b$, based on these scores. Here, the proportional distributor of \cite{klamkin2024bucketized} is adopted:
{\setlength{\abovedisplayskip}{2pt}
  \setlength{\belowdisplayskip}{1pt}\[
\eta(s_b, b) = \lfloor{\beta \frac{s_b}{\sum_{b \in \mathcal{B}} s_b}}\rfloor.
\]}
\subsection{Active Learning Filter}\label{subsec:Active Set Filter}
Line 12 starts the active learning procedure by constructing a filter $\mathcal{X}_b$ for each bucket, $b$, based on the performance of $\hat h$ on validation data in $b({\cal D}_v)$. This filter is defined with a selection function $\sigma$, that balances exploration with exploitation, where
{\setlength{\abovedisplayskip}{3pt}
  \setlength{\belowdisplayskip}{3pt}\begin{align}
\sigma(b(\mathcal{D}_v) = \sigma^{greed}(b(\mathcal{D}_v),\psi n_b)\;\bigcup \;\sigma^{rand}(b(\mathcal{D}_v),(1-\psi) n_b).
\end{align}}
Function $\sigma^{greed}$ performs the exploitation step, selecting the worst performing points in $b(\mathcal{D}_v)$ using 
validation loss function $\mathbb{L}(x,\hat{h})$. Similarly, $\sigma^{rand}$ performs the exploration step, by selecting points uniformly at random from $b(\mathcal{D}_v)$. The split between the two selections is controlled by parameter $\psi \in [0,1]$, where $\psi n_b$ samples are selected from  $\sigma^{greed}$ and $(1-\psi) n_b$ samples are selected from $\sigma^{rand}$.

\subsection{Active Learning Candidate Pool}\label{subsec:Active Learning Candidate Pool}

In general, the high loss.\, as captured by  validation points included in the filter ${\mathcal X}_b$, are a surrogate indicator of underrepresented regions in training data. Line 14 defines a region around each point $x \in {\mathcal X}_b$ that is near to $x$, as defined as sharing a same active-set polytope $\Upsilon_x$. This region defines a candidate pool of training data that shares the same underlying optimization structure as the poor performing validation point.

However, since the active set is unknown for points in $\mathcal{D}_p$, $\Upsilon_x$ is also unknown and therefore requires approximation. To approximate $\Upsilon_x$, a $k$-Nearest Neighbors (kNN) algorithm is applied to a finite subset of $\mathcal{D}_p$. Because of the finiteness, an exact match of the active set for $x$ may not exist, so kNN provides the approximatation estimate. 

Specifically, for each point $x \in \mathcal{X}_b$, the candidate region is defined with 
{\setlength{\abovedisplayskip}{0pt}
  \setlength{\belowdisplayskip}{2pt}\begin{align}
    \hat{\Upsilon}_{x} =\text{kNN}(x, b(\mathcal{D}_p), \hat{h}_a, k_a).
\end{align}} 
where $k_a$ is the number of nearest neighbors (as defined by hamming distance $\delta_H(\tilde{a}^1, \tilde{a}^2) = \sum_{k=1}^{K} \mathbf{1}[\tilde{a}^1_k \neq \tilde{a}^2_k]$ on binary representation of active constraints) selected, and ${\hat h}_a$ is the DNN used to predict the active constraints for points in $b(\mathcal{D}_p)$.

\subsection{Active Learning Selection}\label{subsec:Active Learning Selection}

Line 16 of algorithm \ref{Active Learning Loop} performs the final step of the active sampling with the selection of the data to add to the training set, $\cal D$.  This point is selected by identifying the data from the region that shares the same active set as $x$ ($\Upsilon_x$) that is most similar in other structural properties, as defined by $\delta$.

Here, two choices of $\delta$ are considered in this paper.  The first is defined by the L1-norm in $\mathcal{X}$,
{\setlength{\abovedisplayskip}{1pt}
  \setlength{\belowdisplayskip}{3pt}\begin{align}\label{eqn:AS_raw}
\delta(x,x') = |x - x'|_1.
\end{align}}
and is referred to as \emph{raw} in the experimental results. The second is calculated using the embedded distance in $\mathcal{X}_{pen}$, where $\mathcal{X}_{pen}$ is the space of penultimate layer outputs of the optimization proxy $\hat{h}$.
{\setlength{\abovedisplayskip}{1pt}
  \setlength{\belowdisplayskip}{3pt}\begin{align}\label{eqn:AS_pen}
\delta(x,x') = |\hat{h}^{L-1}(x) - \hat{h}^{L-1}(x')|_1
\end{align}}
and is referred to as \emph{pen} in the experimental results.
The use of penultimate layer outputs as features in place of the original input is common in the active learning literature, particularly for classification problems \cite{kirsch2019batchbald}. 
\begin{figure*}[!t]
    \centering
    \begin{subfigure}{\textwidth}
        \centering
        \includegraphics[width=\linewidth]{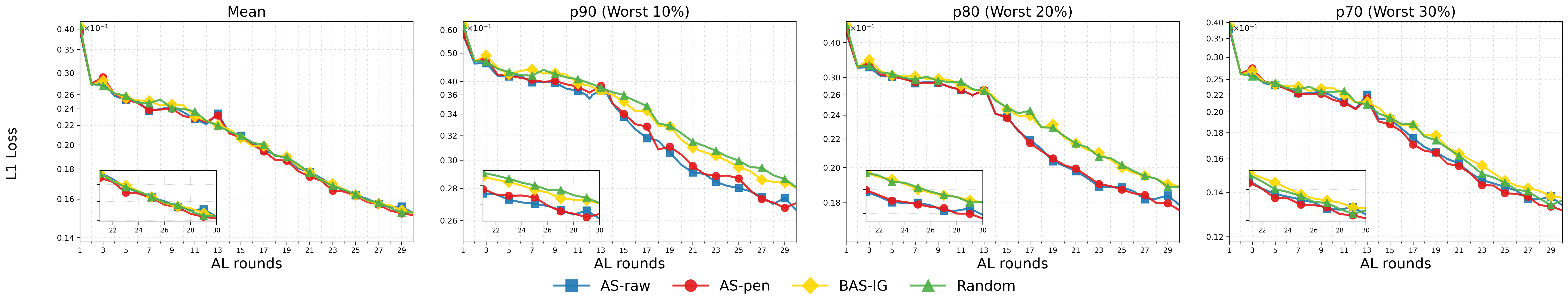}
        \caption{89\_pegase}
        \vspace{-0.05cm}
        \label{fig:sub1}
    \end{subfigure}
    \begin{subfigure}{\textwidth}
        \centering
        \includegraphics[width=\linewidth]{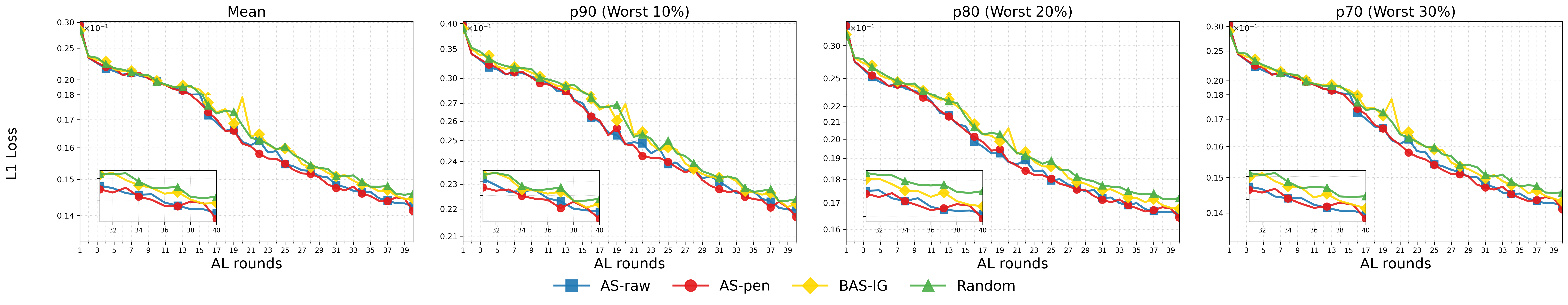}
        \caption{300\_ieee}
        \vspace{-0.05cm}
        \label{fig:sub2}
    \end{subfigure}
    \begin{subfigure}{\textwidth}
        \centering
        \includegraphics[width=\linewidth]{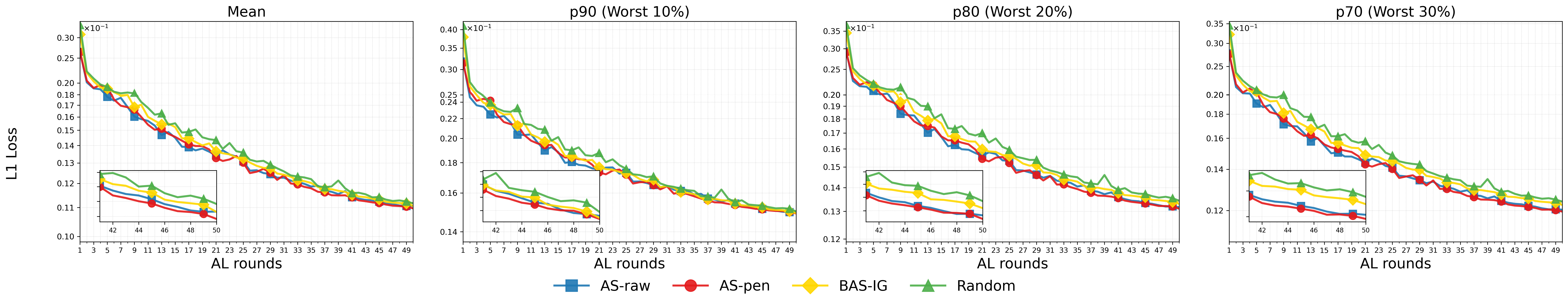}
        \caption{1354\_pegase}
        \vspace{-0.05cm}
        \label{fig:sub3}
    \end{subfigure}
    \caption{A comparison of mean and tail performance}
    \vspace{-0.6cm}
    \label{fig:main_figure}
\end{figure*}
\section{Active Set Illustration}
\begin{figure}[ht!] 
\centering 
\includegraphics[width=0.28\textwidth]{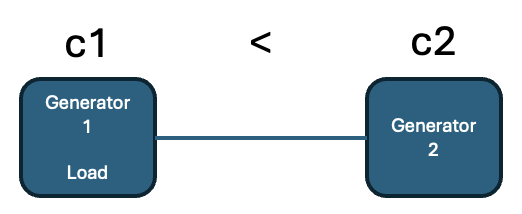} 
\caption{A motivating example} 
\vspace{-0.6cm}
\label{fig:twobus} 
\end{figure}
Before discussing the detailed experimental results, it is useful to illustrate, with a simple example, the role active sets play in training an accurate optimization proxy. In considering baseline algorithms, including \cite{klamkin2024bucketized}, the conjecture that motivated this work was the observation that many of the poor-performing cases corresponded to instances with underrepresented active constraint sets in the training data. Our conjecture suggested a violation of the principal behind successful optimization proxies--similar problem instances always yield similar (predictable) solutions. Instead, we observed that active set transitions induce abrupt changes in the geometry of the solution space that were difficult to predict without sufficient examples in the training data.
To illustrate this phenomena, consider a simple two-bus OPF system (Figure \ref{fig:twobus}) with a single load at bus~1 and generators at each bus. In this example generator~1 has significantly cheaper cost than generator~2, i.e., $c_2 \gg c_1$. In the uncongested regime, where $S_1^d \le \bar{S}_1^g$, all demand is served by generator 1, and an optimization proxy essentially predicts a linear response to changes in $S_1^d$. However, when demand is greater than  $\bar{S}_1^g$, the response is abruptly different.

For this exaggerated illustration, we trained an optimization proxy for a single bucket where the bulk of the load distribution is $\le \bar{S}_1^g$.  In this example $c_1 = 10, c_2 = 30, \bar{S}_1^g = 40$ and the active samples primarily include examples where $S_1^d \le \bar{S}_1^g$. Thus, the optimization is very accurate in this regime, for example, when $S_1^d = 30.16$, the model predicts 302.12, very close to the actual objective of 301.6.  In contrast, when $S_1^d = 40.51$, the model predicts 415.32, which corresponds to a 2.8\% error from the actual objective value of 403.53. However, when our active learning approach is used and includes examples when the constraint on $\bar{S}_1^g$ is active, the optimization proxy predicts 408.59, reducing the error by 43\%. While this example is contrived, it highlights how the geometry of the underlying optimization problem contributes to the success or failure of the optimization proxy.
\section{Numerical Experiments}\label{sec:exp}
Experiments are performed on an NVIDIA L40S GPU. The nonlinear programming solver Ipopt \cite{biegler2009large} is used to compute ACOPF solution data sets.
\subsection{Setup}
The experimental results are presented for public benchmark power networks 89\_pegase, 300\_ieee and 1354\_pegase from PGLib \cite{klamkin2025pglearn}. The proxies $\hat{h}$ and $\hat{h}_a$ are  fully connected DNNs with 3 hidden layers, connected by softplus activations, and optimized with Adam \cite{kingma2014adam}. 2000 samples are labeled to create the initial training set, and in each AL round 300 samples are selected $(\beta)$. The results are reported using a held-out test set after each AL round, and are averaged over 10 trials using different random seeds.

The evaluation uses Algorithm 1 to generate $\mathcal{D}_v$ and $\mathcal{D}_0$ with $\mathcal{Q}_r = \text{Uniform}(0.8, 1.2)$ and $\mathcal{Q}_i = \text{LogNormal}(0, 0.15)$. The unlabeled pool $\mathcal{D}_p$ is generated using the same procedure (Algorithm~1), but the computationally expensive solver step (line~8) is omitted; these instances are labeled on-demand during the active learning process. The evaluation uses $|\mathcal{D}_v| = 2000$, $k=30$, and $b=300$. Two variants of the proposed active set integrated sampling are tested: AS\_raw and AS\_pen. For the AS\_raw variant, the distance $\delta$ is computed directly in the input space $\mathcal{X}$ as defined in \eqref{eqn:AS_raw}. For the AS\_pen variant, the  distance is computed between from the penultimate layer of the proxy model $\hat{h}^{L-1}$ as defined in \eqref{eqn:AS_pen}. The hyperparameter $\psi$ is selected as 0.4.

The method is evaluated against the baselines
\begin{enumerate}
    \item BAS-IG \cite{klamkin2024bucketized}: The bucketization of BAS-IG is based on the load perturbation factor, and applies the input gradient loss as described in \eqref{eqn:acquisition}.
    \item Random: a standard baseline active method that
selects samples uniformly from the input domain. 
\end{enumerate}
\subsection{Numerical Results}
\begin{figure*}[!t]
    \centering
\includegraphics[width=0.9\linewidth]{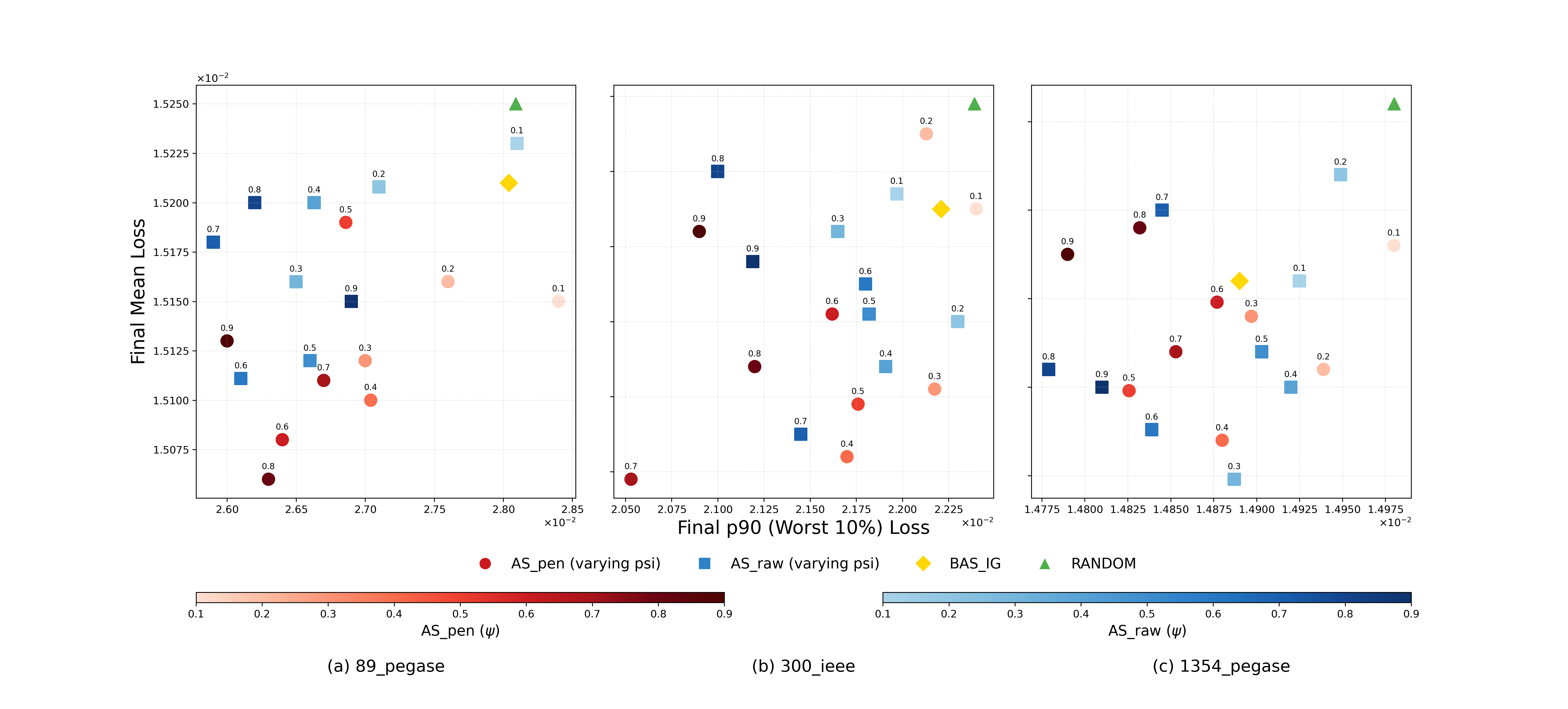}
    \caption{Sensitivity Analysis of the ratio parameter $\psi$}
    \label{fig:sensitivity_figure}
    \vspace{-0.4cm}
\end{figure*}
Figure \ref{fig:main_figure} reports prediction accuracy (L1 loss) measured across AL rounds. Each row corresponds to a benchmark dataset, and columns summarize mean and tail percentiles, which assess robustness: consistently low losses in the tails indicate stronger reliability. Across datasets, AS-pen and AS-raw deliver state-of-the-art performance over all metrics.

Random sampling performs worst on tail metrics because it provides no coverage guarantees; without intentional diversity, performance fluctuates widely in underrepresented regions. BAS methods achieve intermediate tail robustness. A plausible explanation is the adapted BAS bucketization enforces fairness only at the bucket level; overlapping buckets can create imbalance. For example, if an underperforming region appears in multiple validation buckets, it may be overcompensated and oversampled simply because it is counted multiple times.

The AS-pen and AS-raw methods are superior overall because they partition the space into non-overlapping regions and explicitly target weak spots associated with active-set shifts, which are information-rich and drive large errors. Both methods also exhibit the fastest early-stage convergence across metrics—a valuable trait given the tight computational budgets and dynamic data streams characteristic of real-world ACOPF applications. The practical challenge in training complex proxy models is not merely achieving high accuracy, but doing so efficiently, as later active learning training rounds often yield diminishing returns. The central goal of active learning is therefore sample efficiency: accelerating progress toward a near-optimal state under a fixed budget. 

{Validating a key goal of this work, the active learning strategies (AS-pen and AS-raw) deliver superior sample efficiency for tail-risk mitigation, as evidenced by their strong performance on the 70th, 80th, and 90th error percentiles. This advantage is most pronounced in the 89\_pegase case, where reaching 90\% of final converged performance four rounds early saves \emph{at least 1,200} expensive labeling queries. This efficiency holds on the larger 300\_ieee and 1354\_pegase systems, with a consistent speedup of at least two rounds that saves \emph{at least 600} samples, thereby demonstrating the framework's capacity for robust performance under tight budgets.}

Although their overall performance is similar, AS-pen outperforms AS-raw slightly at convergence when applied to large datasets (e.g 1354\_pegase). This aligns with the principle that penultimate-layer representations provide a cleaner, more task-aligned signal than raw inputs, an advantage that is particularly pronounced in high-dimensional spaces where raw features can be noisy.

Moreover, a sensitivity analysis of the hyperparameter $\psi$, which balances exploration and exploitation, was conducted at the final active learning round. Figure~\ref{fig:sensitivity_figure} illustrates this tradeoff by plotting the mean L1 loss (y-axis) against the 90th percentile L1 loss (x-axis) for both AS\_raw and AS\_pen. The parameter $\psi$ is varied from 0.1 to 0.9, with darker colors corresponding to larger values.

The results clearly show that increasing $\psi$ (darker dots) consistently shifts the performance to the left, indicating a lower 90th percentile error. This reveals a practical application: for scenarios where {minimizing tail risk is prioritized over average performance, a larger $\psi$ can be selected to achieve greater robustness}. Conversely, the best mean performance is achieved with mid-range $\psi$ values, demonstrating that a balance between exploration and exploitation is optimal for minimizing average error.

\section{Conclusion and Future work}
This paper introduced a novel active learning framework for ACOPF optimization proxies that leverages the underlying structure of the optimization problem. By using active sets to define a feature space, the framework moves beyond generic sampling techniques to intelligently explore the input domain. The experimental results demonstrate that this approach not only converges faster than state-of-the-art methods but also produces a significantly more robust proxy. Critically, the proposed method shows a marked improvement in reducing tail prediction errors, addressing a key limitation of existing models and advancing the development of trustworthy proxies for real-world grid operations.

This work opens several avenues for future research. One promising direction involves incorporating richer domain-specific knowledge into the sampling process, which could further enhance sample efficiency and target operationally critical scenarios. Another area for exploration is extending and validating the framework on a broader class of complex power system problems, testing the limits of its scalability and generalization to different network topologies and conditions.
\section*{Acknowledgments}
We gratefully acknowledge the discussions with Brian Bell which greatly enhanced the development of this work. The work was partially funded by Los Alamos National Laboratory's Directed Research and Development project, ``Artificial Intelligence for Mission (ArtIMis)'' under U.S. DOE Contract No. DE-AC52-06NA25396.
The work was also partially funded based upon work supported by the National Science Foundation under Grant No. 2112533 and Grant No. DGE-2039655. Any opinions, findings, and conclusions or recommendations expressed in this material are those of the author(s) and do not necessarily reflect the views of the National Science Foundation.
\bibliographystyle{IEEEtran}
\bibliography{ref}
\end{document}